%% file: main.tex
\documentclass[letterpaper, 10 pt, journal, twoside]{ieeetran}

\IEEEoverridecommandlockouts 

\usepackage{graphics}
\usepackage{epsfig}
\usepackage{amsmath}
\usepackage{amsfonts}
\usepackage{cite}
\usepackage{color}
\usepackage{multirow}
\usepackage{subcaption}
\usepackage{hyperref}
\usepackage{booktabs}
\usepackage{import}

\input{symbols.tex}

\graphicspath{{./figures/}}

\hypersetup{
    colorlinks=true,
    linkcolor=blue,
    urlcolor=blue,
}

\title{
DeepFactors: Real-Time Probabilistic Dense Monocular SLAM
}

\author{Jan Czarnowski, Tristan Laidlow, Ronald Clark, and Andrew J. Davison%

\thanks{Manuscript received: September, 10, 2019; Revised December, 1, 2019;
Accepted December, 23, 2019.}
\thanks{This paper was recommended for publication by Editor Sven Behnke upon evaluation of the Associate Editor and Reviewers' comments.}
\thanks{Research presented in this paper has been supported by Dyson Technology Ltd.}
\thanks{The authors are with the Dyson Robotics Laboratory, Imperial College London, United Kingdom {\tt\small j.czarnowski15@imperial.ac.uk}}
\thanks{We would like to thank Michael Bloesch for useful insights during the initial stages of the project.}
\thanks{Digital Object Identifier (DOI): see top of this page.}

} 

\begin{document}

\maketitle

\begin{abstract}
    The ability to estimate rich geometry and camera motion from monocular imagery is fundamental to future interactive robotics and augmented reality applications. Different approaches have been proposed that vary in scene geometry representation (sparse landmarks, dense maps), the consistency metric used for optimising the multi-view problem, and the use of learned priors.
    We present a SLAM system that unifies these methods in a probabilistic framework while still maintaining real-time performance. This is achieved through the use of a learned compact depth map representation and reformulating three different types of errors: photometric, reprojection and geometric, which we make use of within standard factor graph software.
    We evaluate our system on trajectory estimation and depth reconstruction on real-world sequences and present various examples of estimated dense geometry.
\end{abstract}


\section{Introduction}

\IEEEPARstart{R}{esearch} in monocular SLAM has been divided into two paradigms that mainly differ in the geometric map representation. \textit{Sparse} methods\cite{Davison:ICCV2003, Klein:Murray:ISMAR2007, Mur-Artal:etal:TRO2015} form and maintain a map consisting of a set of point landmarks which are estimated through observing, identifying and tracking them in the camera images. In order to enable repeatable recognition and reliable matching of landmark observations, the map is limited to a relatively small number of salient points that are characteristic and easily distinguishable in their image projections (typically corners or edges). This limits the usefulness of resulting map reconstructions, which cannot be used in interactive robotics tasks or advanced augmented reality applications. At the same time, the size of the representation allows for real-time joint probabilistic inference. Since sparse methods rely on gradient-based keypoint detectors\cite{Lowe:IJCV2004, Rublee:etal:ICCV2011, Leutenegger:etal:ICCV2011} and reprojection error that is formulated as distance on the image plane, they are robust to image noise, outliers and lighting variations.

With the advent of General Purpose GPU (GPGPU) programming whole-image alignment has become cheap and accessible. This has spawned a new set of methods that use all image pixels and estimate a \textit{dense} and more useful reconstruction of the observed scenes\cite{Newcombe:etal:ICCV2011}. Due to the increased computational demands, dense methods simplify the inference framework by discarding variable cross-correlations and alternating between tracking and mapping. More pixel information used in the estimation process allows for increased robustness to image degradation due to motion blur, but the reliance on photometric error measured in pixel intensity makes dense methods fragile to sequences violating the brightness constancy assumption.

\begin{figure}[t]
    \centering
    \includegraphics[width=\columnwidth]{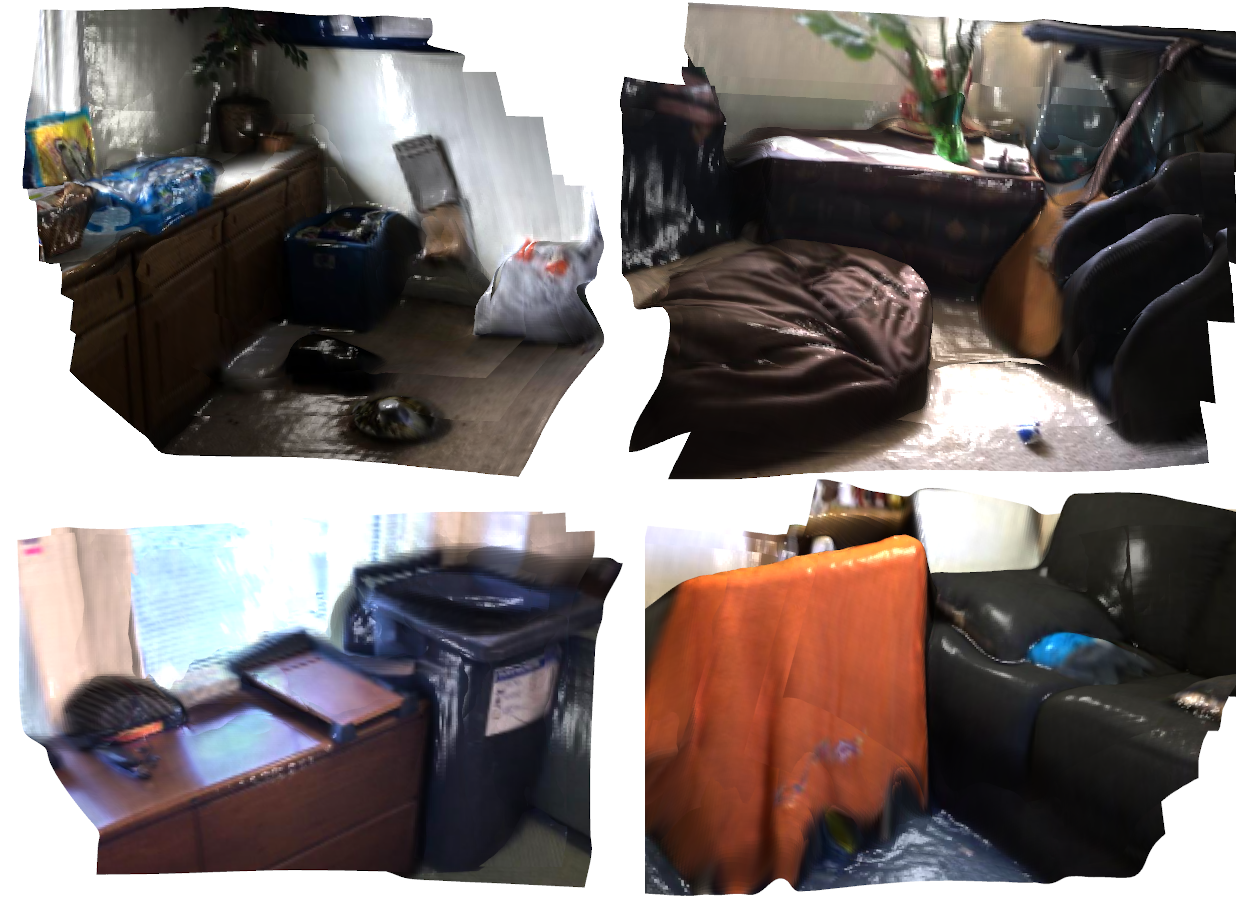}
    \caption{Example reconstructions of scenes from the ScanNet validation set created with our system. Reflective shading has been exaggerated in order to highlight structure.}
    \label{fig:reconstr_highlight}
    \vspace{2mm}\hrule
\vspace{-0.5cm}
\end{figure}

The success of deep learning inspired a variety of new systems. Learned components have been integrated into the SLAM pipeline to a differing degree, with solutions ranging from classical model based systems delegating sub-components to a neural network (e.g.\cite{Weerasekera:etal:ICRA2017, Pizzoli:etal:ICRA2014}) to purely end-to-end learned systems that directly predict the camera poses and the geometry of the scene (e.g.\cite{Clark:etal:AAAI2017, Zhou:etal:CVPR2017, Yin:Shi:CVPR2018}). Priors learned directly from data allow solving difficult tasks such as depth prediction from monocular imagery \cite{Eigen:etal:NIPS2014, Ummenhofer:etal:ARXIV2016, Liu:etal:2015, Garg:etal:ECCV2016}.

We present \textit{DeepFactors}, a real-time SLAM system that builds and maintains a dense reconstruction but allows for probabilistic inference and combines the advantages of different SLAM paradigms. It also presents a tight integration of learning and model based methods through a learned compact dense code representation that drives significant changes to the core mapping/tracking components of the SLAM pipeline. The main contributions presented in this paper can be summarised as follows:
\begin{itemize}
    \item The first real-time probabilistic dense SLAM system,
    \item A system that integrates learned priors over geometry with classical SLAM formulations in a probabilistic factor-graph formulation.
\end{itemize}

\section{Related Work}
In CodeSLAM\cite{Bloesch:etal:CVPR2018}, the authors have introduced the concept of learning a compact optimisable representation and utilising it to solve the dense structure-from-motion problem. The feasibility of this idea has been proved by implementing a windowed 3D reconstruction/visual odometry. The presented system lacked features of full SLAM, was not capable of real-time performance and did not generalise to real handheld camera scenes. 

Our work builds on the CodeSLAM idea and explores the impact of the code optimisation on traditional SLAM pipelines. DeepFactors is a complete new SLAM system build from scratch with a different mapping backend, error formulations (the sparse geometric and reprojection errors) and all the various design choices within the SLAM algorithm like keyframing, map maintenance or tracking. It contains features that CodeSLAM was missing -- local and global loop closure or relocalisation and optimises the full map in batch instead of a fixed window only.The optimised GPU usage, efficient implementation and SLAM design choices enable real-time performance and the use of a standard factor graph software allows for straightforward probabilistic integration of different sensor modalities which was not possible with the previous formulations of dense SLAM.

An example of other directly comparable work is CNN-SLAM\cite{Tateno:etal:CVPR2017}. Although with a substantially different principle of operation, the authors present a real-time full SLAM system that builds a large scale map and supports loop closures by utilising LSD-SLAM \cite{Engel:etal:ECCV2014} and incorporating learned priors. We use CNN-SLAM as a baseline comparison for our method.

There exists a range of systems with learned components which are not fully-featured SLAM systems but instead focus on multi-frame dense reconstruction\cite{Laidlow:etal:ICRA2019, DBLP:journals/corr/abs-1809-02966, DBLP:journals/corr/abs-1901-02571, Weerasekera:etal:ICRA2017}. In BA-Net\cite{DBLP:journals/corr/abs-1806-04807} the authors use a technique similar to CodeSLAM where a set of basis depth maps is predicted from the image and optimised in a bundle adjustment problem to find a dense per-pixel reconstruction. The system has been trained end-to-end with the optimisation included which might result in a learned representation better suited for the structure from motion problem. In contrast to our work, BA-Net is not a real-time SLAM system that builds and optimises a consistent multi-keyframe map.

A notable mention is DeepTAM\cite{Zhou:etal:ECCV2018}, which builds upon DTAM\cite{Newcombe:etal:ICCV2011} by replacing both the TV-L1 optimisation and camera tracking with a deep convolutional neural network and achieves results outperforming standard model-based methods. In contrast to our work, it follows the same tracking and mapping split used in all dense methods and is not capable of real-time operation. Although the authors took special care to address the generalisation problem, their system still ultimately relies on seeing all possible variations of input data, which is hard in the case of full 6 DoF motion in real world conditions. Our approach relies less on the network generalisation as we perform optimisation that allows to correct for bad network predictions as in our system, neural networks are mainly used for obtaining an image conditioned manifold to optimise over. 

In contrast to the previously mentioned methods, our work strives towards the goal of a unified SLAM framework that incorporates both learned and model-based methods as well as dense and sparse approaches to localisation and mapping and possibly points towards a new generation of SLAM systems. In the remainder of our paper, we explain the building blocks of our system and show evaluation on real world sequences.

\section{Code Based Optimization}
\label{sec:code_optimisation}
To reconstruct a dense representation of the scene geometry and estimate the camera motion we formulate a multi-view dense bundle adjustment problem. We parametrise the reconstructed geometry $G$ as a set of depth maps at each camera frame $G = \{\matD_0, \matD_1, ..., \matD_n\}$. In a na\"ive formulation, pixels of each depth are uncorrelated and optimised independently, which makes the problem too ill-posed and costly to solve due to the large number of parameters.

\begin{figure}
	\centering
	\includegraphics[width=0.8\linewidth]{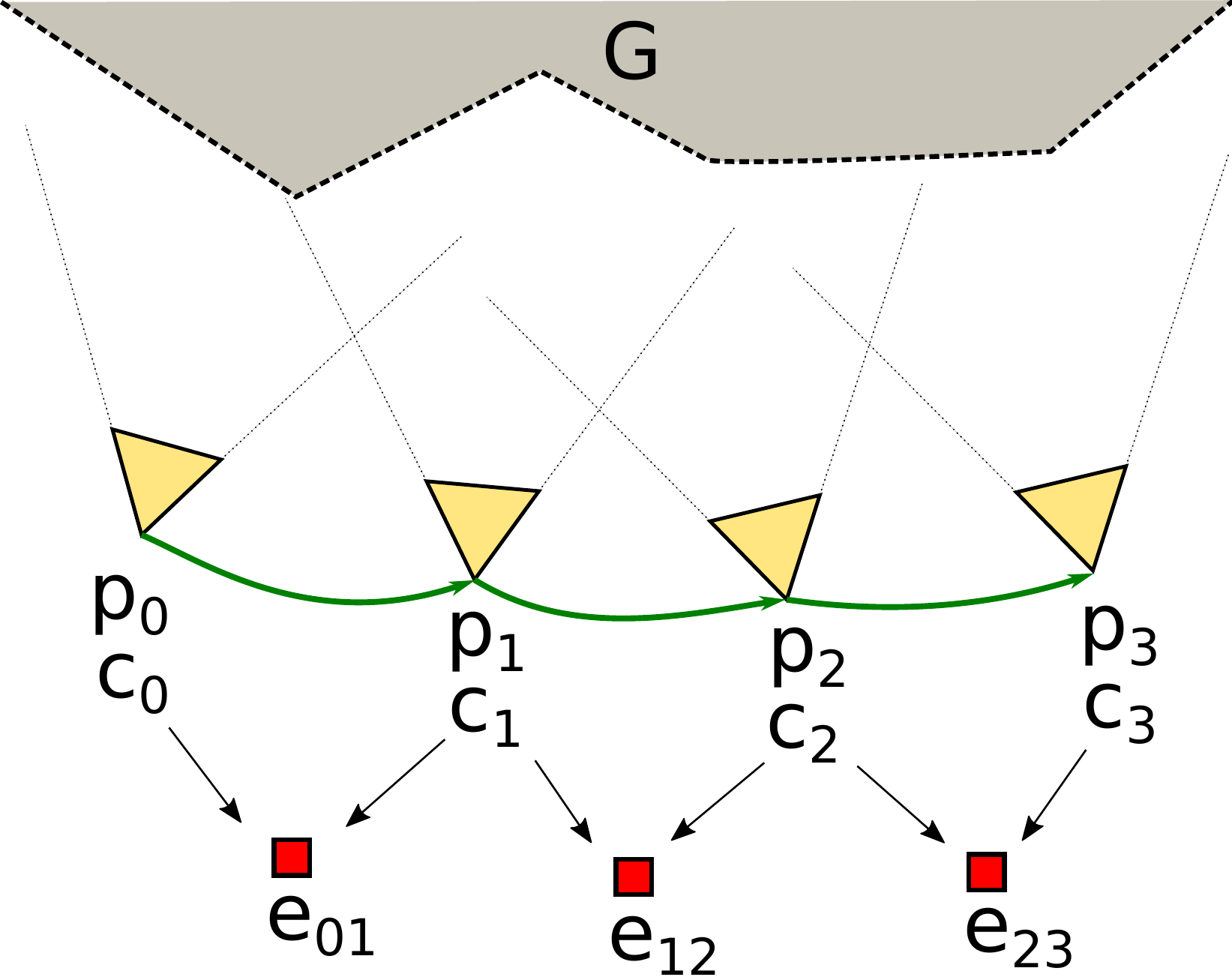}
	\caption{Dense multi-frame Structure From Motion problem using an optimisable  compact dense code representation. Each camera frame $i$ consists of a 6DoF world pose $p_i$ and an a associated code $\vecc_i$. We minimise various pairwise consistency losses $e_{ij}$ in order to find the best estimate for scene geometry $G = \{D_0(c_0), ..., D_n(c_n)\}$ and camera poses $p_0 ... p_N$.}
	\label{fig:code_sfm}
    \vspace{2mm}\hrule
\end{figure}

Following the methodology proposed in~\cite{Bloesch:etal:CVPR2018} we optimise depth on a learned compact manifold (code) to mitigate both of these problems (Figure \ref{fig:code_sfm}). We express the depth map $\matD_i$ of a frame $i$ as a function of code $\vecc_i$ and the associated image $\matI_i$. In order to avoid costly relinearisations during optimisation, we require this relation to be linear:

\begin{equation}
    \matD_i = f(\vecc_i, \matI_i) = \matD_i^0 + \matJ(\matI_i) \vecc_i,
\end{equation}

\noindent
where $\matD_i^0 = f(\mathbf{0}, \matI_i)$ is the depth map resulting from decoding an all-zero code and $\matJ(\matI_i) = \frac{\partial \matD_i}{\partial \vecc_i}$ is the image-conditioned Jacobian. 

\begin{figure}
    \centering
	\centering
	\includegraphics[width=1\linewidth]{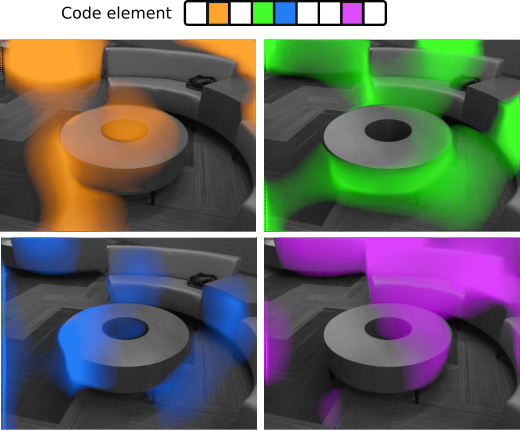}
    \caption{Visual representation of the code Jacobian $\frac{\partial D_i}{\partial \vecc_i}$. Perturbing different code elements affects various regions of the depth image.}
    \label{fig:code_jacobian}
    \vspace{2mm}\hrule
    \vspace{-0.5cm}
\end{figure}

When optimising on the code manifold, groups of depth pixels are correlated together which makes the optimisation problem more tractable. Figure~\ref{fig:code_jacobian} presents the pixels affected by perturbing different elements of the latent code vector.

Using the code representation, we minimise a set of different objective functions that ensure consistency between observations from the overlapping camera frames and find the best estimate of scene geometry and camera motion. These functions are used as pairwise constraints in the factor graph used by our SLAM system. The following sections describe these factors in greater detail.

\subsection{Photometric Factor}
A consistency loss typically used in Dense SLAM is the photometric error. It measures the difference directly between source image intensities $\matI_i$ and the target image $\matI_j$ warped into frame $i$:

\begin{equation}
    e_{pho}^{ij} = \sum_{\vecx \in \Omega_i}  || \matI_i(\vecx) - \matI_j(\omega_{ji}(\vecx, \vecc_i, \matI_i)) ||^2,
\end{equation}

\noindent
where $\omega_{ji}$ warps pixel coordinates $\vecx$ in frame $i$ to frame $j$:
\begin{equation}
    \omega_{ji}(\vecx, \vecc_i, \matI_i) = \pi(\matT_{ji}(\pi^{-1}(\vecx, \matD_i(\vecx)))),
\end{equation}

\noindent
where $\pi$ and $\pi^{-1}$ are the projection and reprojection function respectively, $\matD_i(\vecx) = \matD(\vecx, \vecc_i, \matI_i)$ is the depth map decoded from code $\vecc_i$ and $\matT_{ji} \in SE(3)$ is the relative 6DoF transformation from frame $i$ to $j$.

\subsection{Reprojection Factor}
We also use the indirect reprojection error widely used in classical  structure from motion. Given a set of matched landmark observations between the images, this method measures the differences between their observed and hypothesised locations:

\begin{equation}
    e_{rep}^{ij} = \sum_{(\vecx,\vecy) \in M_{ij}} || \omega_{ji}(\vecx, \vecc_i, \matI_i) - \vecy ||^2,
\end{equation}

\noindent
where $M_{ij}$ is a set of salient image features matched between frame $i$ and $j$. To handle mismatched features, we use Cauchy robust cost function that has a constant response to outliers. We use BRISK~\cite{Leutenegger:etal:ICCV2011} to detect and describe key points in images.

\subsection{Sparse Geometric Factor}
Another form of consistency can be expressed with differences in scene geometry. In our simplified form, we compare depth map $D_j$ with depth map $D_i$ warped into frame $j$:

\begin{equation}
    e_{geo}^{ij} = \sum_{\vecx \in \Omega_i} || [\matT_{ji}(\pi^{-1}(\vecx, \matD_i(\vecx)))]_z - \matD_j(\hat\vecx) ||^2,
\end{equation}

\noindent
where $\hat\vecx = \omega_{ji}(\vecx,\vecc_i,\matI_i)$ and $[\vecx]_z$ denotes taking the $z$ component of the vector $\vecx$. We use Huber norm\cite{Huber:AMS:1964} on the error as a robust cost function. In order to save computation, we evaluate the loss only for a sparse set of uniformly sampled pixels. It is possible to sample a different set of pixels at each iteration to stochastically optimise the loss over the whole image.

\section{Network Architecture}

\begin{figure}
    \centering
    \includegraphics[width=1\linewidth]{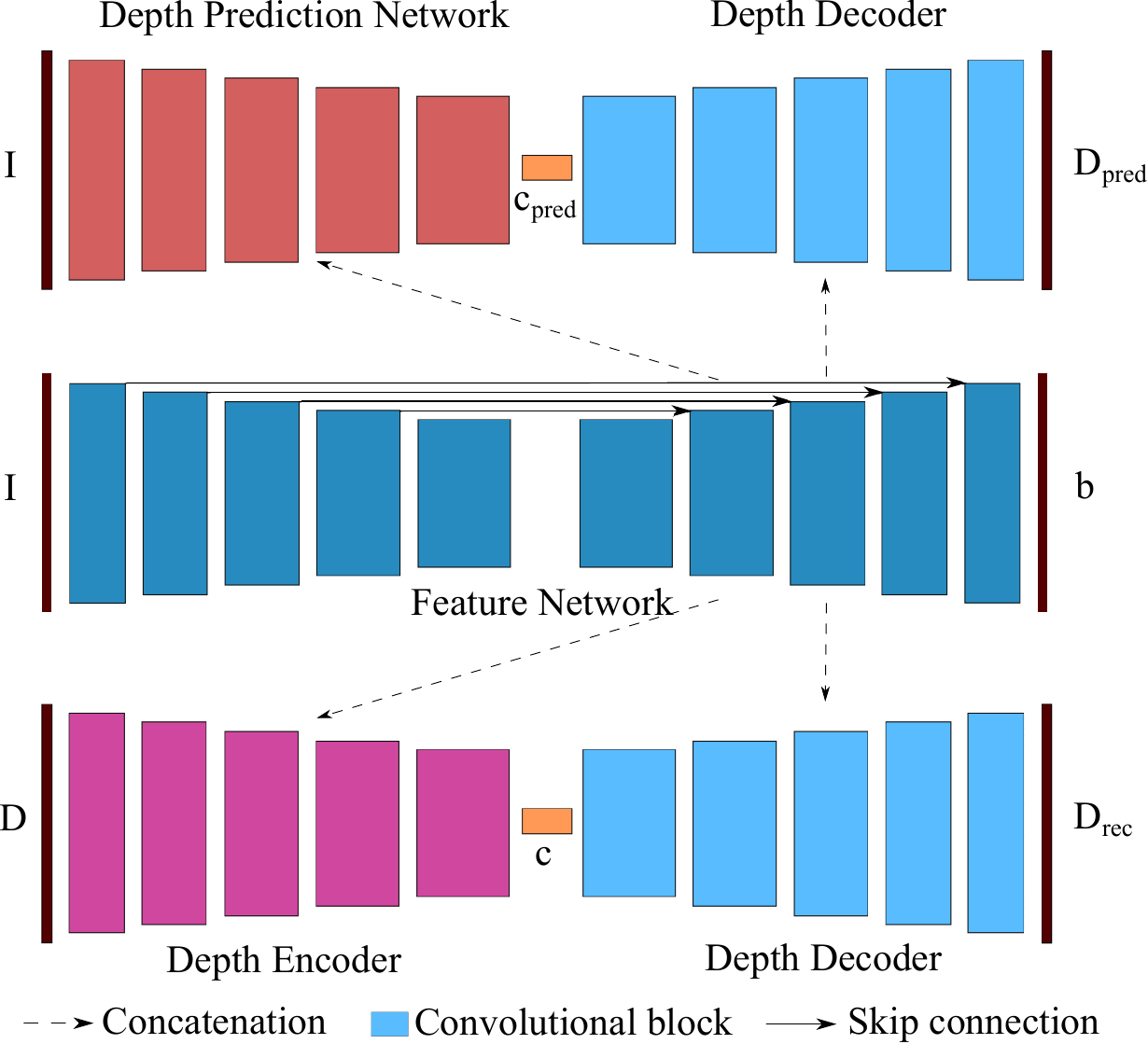}
    \caption{Network architecture. The bottom path is a U-Net depth auto-encoder without skip connections that learns the optimisable compact code $c$. Its encoder and decoder are conditioned on image features concatenated from the Feature Network. The top part of the network learns to predict optimal code $c_{pred}$ from the input RGB image that is decoded into a mono depth prediction. The depth decoder is shared between the two networks (light blue). }
    \label{fig:network_architecture}
    \vspace{2mm}\hrule
\end{figure}

For learning the compact code representation we modify the network from CodeSLAM\cite{Bloesch:etal:CVPR2018}. The full network architecture is presented in Figure  \ref{fig:network_architecture}. The middle part represents a U-Net \cite{Ronneberger:etal:MICCAI2015} extracting features from the input RGB image. The input is processed with blocks of convolutions with each block reducing the size and applying multiple convolutions on the reduced resolution. The bottom part of the figure depicts the main Variational Auto-Encoder(VAE)\cite{Kingma:Welling:ICLR2014} that learns the optimisable compact depth representation. The encoder and decoder are conditioned on the above-mentioned features using concatenations.

Similar to CodeSLAM, in order to keep the relation between the reconstructed depth $D_{rec}$ and the code $c$ linear, we do not use any non-linear activations in the depth decoder. To also ensure that the input image retains influence on that relation, we add an element-wise multiplication of each two concatenated layers in the conditioning concatenations.

\begin{figure*}
\centering
\includegraphics[width=\textwidth]{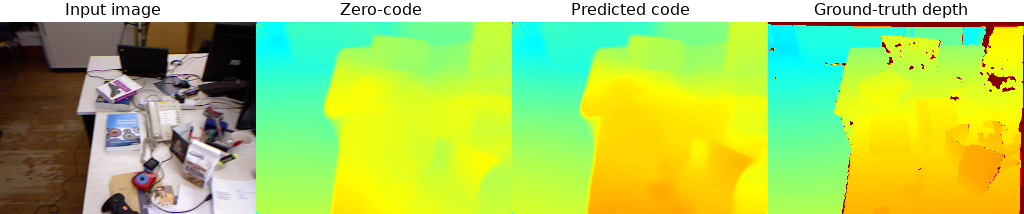}
\caption{A comparison of two ways of initialising new depth maps given an input intensity image: zero-code and explicit code prediction. The inclusion of an additional network that predicts an initial code directly from the image allows for better initial depth estimates. }
\vspace{2mm}\hrule
\label{fig:code_prediction}
\end{figure*}

Due to the KL-divergence based latent loss applied to the code of the depth VAE, an all-zero code corresponds to a likely depth map for the input image $\matI$. In our experiments with running the system on a real camera, we have found that we can achieve better initial depth predictions by augmenting the network with a separate encoder that explicitly predicts an optimal code from the input image $\matI$. For the predicted code to lie in the same space as the learned code $c$, we apply the same latent loss to it.
The added explicit network path is less constrained than the zero code prediction, which allows to achieve better results, as shown in Figure \ref{fig:code_prediction}.

The network also predicts an uncertainty parameter $b$ with the Feature Network which is used in a multi-resolution negative log of Laplacian likelihood loss for the reconstructed depth $D_{rec}$:

\begin{equation*}
    \sum_{x\in\Omega}\frac{|\matD_{rec}(x)-\matD(x)|}{b(x)} + \log(b(x)),
\end{equation*}

\noindent
where $\Omega$ is the set of all pixel coordinates of the input depth $D$. The predicted depth map is supervised with an L1 loss:

\begin{equation}
    \sum_{x\in\Omega} |\matD_{pred}(x) - \matD(x)|.
\end{equation}

\section{System}

The system builds and maintains a keyframe map. Incoming new camera images are resized and corrected to match the network focal length and tracked against the nearest keyframe (section~\ref{sec:tracking}). Once sufficient baseline and other criteria are met, a new keyframe is initialised at the estimated pose with an initial code prediction and added to the graph (section ~\ref{sec:kf_init}). 

To optimise the map, we maintain a factor graph of the batch MAP problem and optimise it each time new observations are introduced. Each new keyframe is connected to last N keyframes in the map using selected pairwise consistency factors presented in section~\ref{sec:code_optimisation}. Real-time performance of solving for the batch solution is achieved by using an incremental mapping algorithm (section~\ref{sec:mapping}).

To increase performance, the system alternates between tracking and joint mapping. After creating a new keyframe, the graph is being optimised for a set number of iterations or until convergence. During that time the tracking and mapping optimisation steps are interleaved to ensure that the system keeps up with incoming new camera images.

To obtain good quality photometric signal a mixture of low and high baseline image pairs is required. Since we only connect last N keyframes with pairwise constraints, this might not always be the case. To mitigate this while keeping the computational complexity low we introduce "one-way" frames that do not have attached depth and are used to feed information to refine the latest keyframe. After optimising for a set amount of steps, active one-way frames are marginalised and removed from the graph. This allows for inexpensive integration of many views into the optimisation and quality reconstruction of depth.

The details of each component of the system have been described in the following sections.

\subsection{Camera Tracking}
\label{sec:tracking}
Each camera frame is tracked against the closest keyframe using our GPGPU implementation of a standard direct whole-image SE3 Lucas-Kanade~\cite{Lucas:Kanade:IJCAI1981, Newcombe:PHD2012, Lovegrove:PHD2011}. In case tracking is lost, we perform relocalization by attempting to align a small resolution image against all keyframes. 

\subsection{Incremental Keyframe Mapping}
\label{sec:mapping}
\begin{figure}
    \centering
    \includegraphics[width=0.8\linewidth]{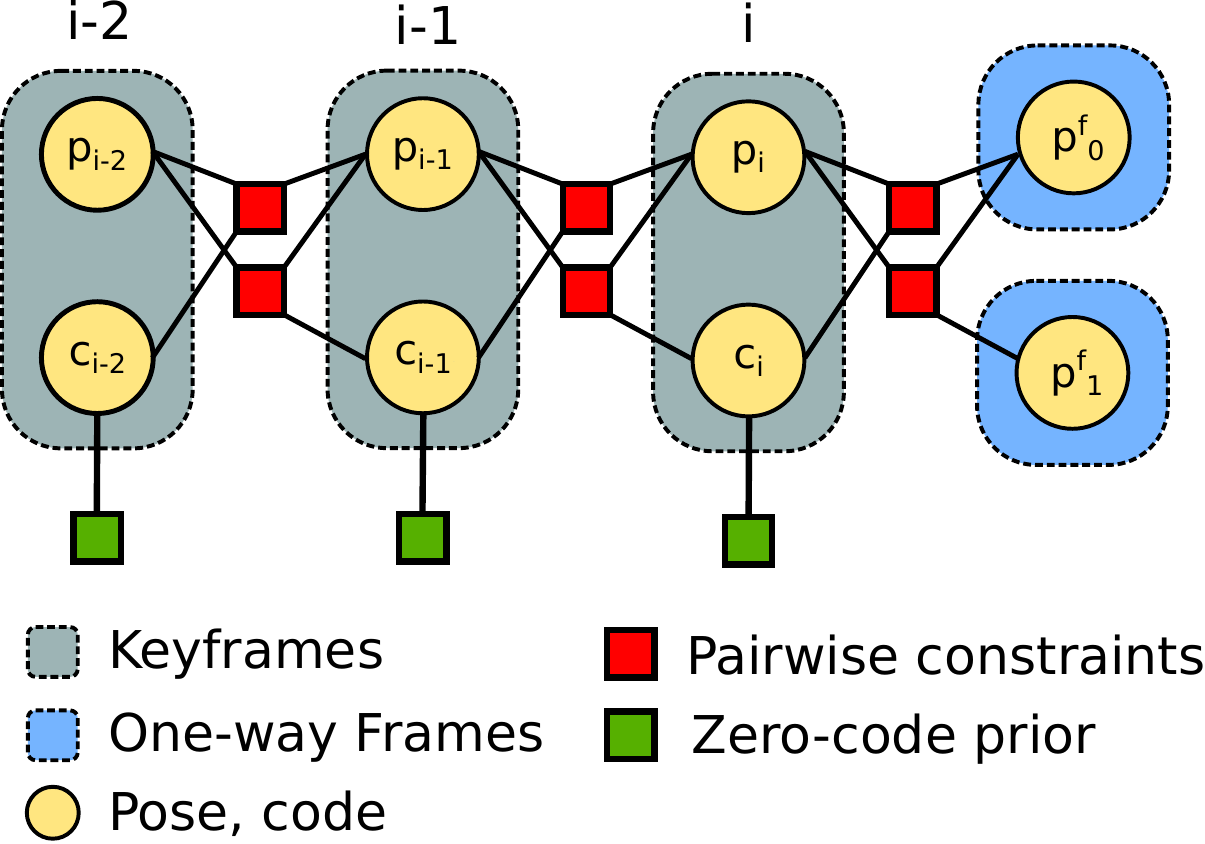}
    \caption{An example instance of the factor graph used in our system containing three keyframes and two one-way frames attached to the latest keyframe. Multiple different pairwise constraints can be used simultaneously (photometric, geometric or reprojection) but for simplicity, only a single type has been presented in this figure.}
    \label{fig:factor_graph}
    \vspace{2mm}\hrule
\end{figure}

We formulate and jointly optimise a batch MAP estimation problem involving all keyframes in the map. Figure \ref{fig:factor_graph} presents a factor graph representation of an example instance of a map built by our system. Each keyframe is represented by a pose $p_i$ and a code $c_i$ variable with variables of neighbouring keyframes involved in pairwise consistency factors (photometric, reprojection or geometric factors). Since the factors were designed to represent a single-way warping between two keyframes, they allow optimisation of the code/depth of only a single keyframe of the pair. Two factors are added to optimise both keyframes. Because during training the code manifold is enforced to be close to a zero-mean Gaussian (variational latent loss), the code has to be kept within the appropriate region during optimisation by using zero-code prior factors that regularise it.

The mapping step performs batch optimisation of all keyframes in the map using standard factor graph software \cite{Dellaert:TechReport2012}. To make optimisation feasible in large scale scenarios we rely on an incremental mapping algorithm -- iSAM2\cite{Kaess:etal:IJRR2012}. It stores a factorisation of the batch Jacobian in the form of a Bayes Tree and incrementally updates it when new variables are added. Only the affected factors are relinearised and re-factorised, which greatly limits the computation required for obtaining the batch solution and allows near constant time updates in an exploration-type movement.

In order to enable marginalisation of one-way frames, we enforce a specific elimination order of the graph variables to ensure that they are leaves in the clique tree built and maintained by iSAM2.

\subsection{Initialization}
\label{sec:kf_init}
We use the explicit code prediction network to obtain an initial code for each new keyframe. The depth decoder is then used to transform the codes into a depth map estimate. On start, the system can be trivially initialised in the exact same manner. The network prediction must be good enough for tracking the initial camera movement so that new views can be fed into the system to refine the depth prediction.

In cases when the single image prediction fails a multi-frame initialisation can be used, in which a keyframe is created for each input frame and the initial graph is optimised before starting the system. We found the single frame initialisation to perform well enough for most scenes.

\subsection{Loop Closure}
We detect and close local and global loops. Within an active region of last 10 keyframes, we assume the relative estimated poses to be correct and therefore use a pose-based criteria to detect loops and add additional pairwise constraints between keyframes. This tightens the graph and allows for more consistent local reconstructions.

At each new input frame we also test for global loop closure events. Outside of the active window we assume that too much drift has been accumulated and therefore we cannot use our estimates to close loops. Initial loop candidates are detected using a bag of words approach\cite{GalvezTRO12} and later further eliminated by attempting to track the current frame against each of them and checking the resulting number of inliers and the estimated pose distance. Because of the photometric error typically having a smaller convergence basin we add only reprojection factors when closing a detected global loop between two keyframes.

\section{Experimental Results}
\subsection{Training}
We have trained our network on a fragment of the ScanNet dataset\cite{Dai:etal:CVPR2017} following the official training/test split. The depth data comes from a real sensor and therefore contains missing values. Since the dataset provides PLY models of the full scene reconstructions together with ground truth poses, we have rendered depth maps from the models and combined them with the raw sensor data to obtain final merged depth maps. We used around 1.4M images in total.

The network was trained for 13 epochs with learning rate $0.0001$, image size $256\times192$ and code size 32. This is the maximum code size that we can achieve real-time results with in our SLAM system.

\subsection{Ablation Studies}

\textit{DeepFactors} combines three different error metrics/factors -- photometric, geometric and reprojection to estimate the camera trajectory and the observed scene geometry. In order to determine how each factor type influences the system performance, we have evaluated various configurations on the quality of the estimated trajectory and reconstruction. The photometric factor is treated as a baseline system and the other two types are included both individually and together. We perform this evaluation on selected shortened scenes from the validation set of the ScanNet dataset and use three error metrics: RMSE of the Absolute Trajectory Error (ATE-RMSE), the absolute relative depth difference (absrel)\cite{Eigen:etal:NIPS2014} and the average percentage of pixels in which the estimated depth falls within 10\% of the true value (pc110)\cite{Tateno:etal:CVPR2017}. The results are presented in Table~\ref{tab:factor_ablation}.

The inclusion of either of the factors reduces the trajectory error and increases the reconstruction accuracy, with the reprojection factor typically having a stronger influence on the former and the geometric factor on the latter. Explicit feature matching utilised within the reprojection error improves local minima avoidance and increases convergence rate, which adds robustness to the system. The geometric error introduces a prior about the world that only a single surface is observed and pins separate depth maps together to form a single reconstruction in the textureless areas that lack photometric information. Combining all three factors achieves the best trajectory and reconstruction results.

\begin{table}
\centering
\def\arraystretch{0.9}
\begin{tabular}{l l l l l}
    \toprule
    \multicolumn{5}{c}{\textbf{Factor Comparison}} \\
    \midrule
    \textbf{Sequence} & \textbf{Factors Used} & \textbf{ATE RMSE$\downarrow$} & \textbf{absrel$\downarrow$} & \textbf{pc110$\uparrow$} \\
    \midrule
    \multirow{4}{*}{scene0565\_00}  & pho      & 0.128 & 0.108 & 57.56\% \\
                                    & pho+rep  & \textbf{0.112} & 0.104 & 59.80\% \\
                                    & pho+geo  & 0.115 & 0.103 & 59.75\% \\
                                    & combined & 0.114 & \textbf{0.102} & \textbf{60.13\%} \\
    \midrule
    \multirow{4}{*}{scene0084\_00}  & pho      & 0.131 & 0.085 & 69.14\% \\
                                    & pho+rep  & 0.074 & 0.082 & 71.05\% \\
                                    & pho+geo  & 0.120 & 0.081 & 71.81\% \\
                                    & combined & \textbf{0.061} & \textbf{0.077} & \textbf{73.66\%} \\
    \midrule
    \multirow{4}{*}{scene0606\_02}  & pho      & 0.089 & 0.214 & 37.22\% \\
                                    & pho+rep  & 0.071 & 0.201 & 39.39\% \\
                                    & pho+geo  & 0.067 & 0.168 & 44.45\% \\
                                    & combined & \textbf{0.066} & \textbf{0.162} & \textbf{46.16\%} \\
    \bottomrule
\end{tabular}
\caption{Comparison of the influence of different factor types on the estimated trajectory and reconstruction errors (\textit{pho:} photometric. \textit{geo:} geometric, \textit{rep:} reprojection)}.
\label{tab:factor_ablation}
\vspace{2mm}\hrule
\end{table}

\subsection{Reconstruction}
We evaluate our reconstruction accuracy against CNN-SLAM as it is the only relevant system that evaluates reconstruction using the whole estimated trajectory and without using the ground-truth poses. Since the authors do not provide implementation of their system, we follow their evaluation strategy by taking the results reported in their paper and using the same sequences from the ICL-NUIM\cite{Handa:etal:ICRA2014} and TUM\cite{Sturm:etal:IROS2012} datasets. For all the keyframes produced by each system, we their estimated depth against the ground-truth by calculating the percentage of the pixels for which the depth is within 10\% of the true value. The results are presented in Table \ref{fig:reconst_eval}.

Since our system is monocular and does not produce up-to-scale trajectories and reconstructions, we use the optimal scale calculated with the TUM benchmark scripts to scale both the trajectory and the depth maps (as done in e.g. \cite{Zhou:etal:CVPR2017}). 

We outperform all compared methods on most of the sequences and on the average. Our system performs worse on the tum/seq2 trajectory (\textit{fr3\_nostructure\_texture\_near\_withloop}) as the camera observes a flat textured wall and due to the small code size (32) used in our system we are typically not able to represent fully flat depth easily.

\begin{table}
\centering
\def\arraystretch{1.1}
\begin{tabular}{p{1.3cm} p{1cm} p{1.5cm} p{1cm} p{1cm}}
    \toprule
    \multicolumn{5}{c}{\textbf{Perc. Correct Depth [\%]}} \\
    \midrule
    Sequence       & Ours            & CNN-SLAM \cite{Tateno:etal:CVPR2017} &
    LSD-BS\cite{Engel:etal:ECCV2014} & 
    Laina\cite{Laina:etal:CoRR2016} \\
    \midrule
    icl/office0    & \textbf{30.17}  & 19.41           & 0.60           & 17.19      \\
    icl/office1    & 20.16           & \textbf{29.15}  & 4.76           & 20.84      \\
    icl/living0    & \textbf{20.44}  & 12.84           & 1.44           & 15.01      \\
    icl/living1    & \textbf{20.86}  & 13.04           & 3.03           & 11.45      \\
    tum/seq1       & \textbf{29.33}  & 12.48           & 3.80           & 12.98      \\
    tum/seq2       & 16.92           & \textbf{24.08}  & 3.97           & 15.41      \\
    tum/seq3       & \textbf{51.85}  & 27.40           & 6.45           & 9.45       \\
    \midrule
    \textbf{Avg.}  & \textbf{27.10}  & 19.77           & 3.44           & 14.62      \\
    \bottomrule
\end{tabular}
\caption{Evaluation of average percentage of pixels for which the estimated depth falls within 10\% of the true value on the ICL-NUIM and TUM datasets (tum/seq1: \textit{fr3\_long\_office\_household}, tum/seq2: \textit{fr3\_nostructure\_texture\_near\_withloop}, tum/seq3: \textit{fr3\_structure\_texture\_far})}
\label{fig:reconst_eval}
\vspace{2mm}\hrule
\end{table}

\begin{figure*}[h]
\centering
\includegraphics[width=\textwidth]{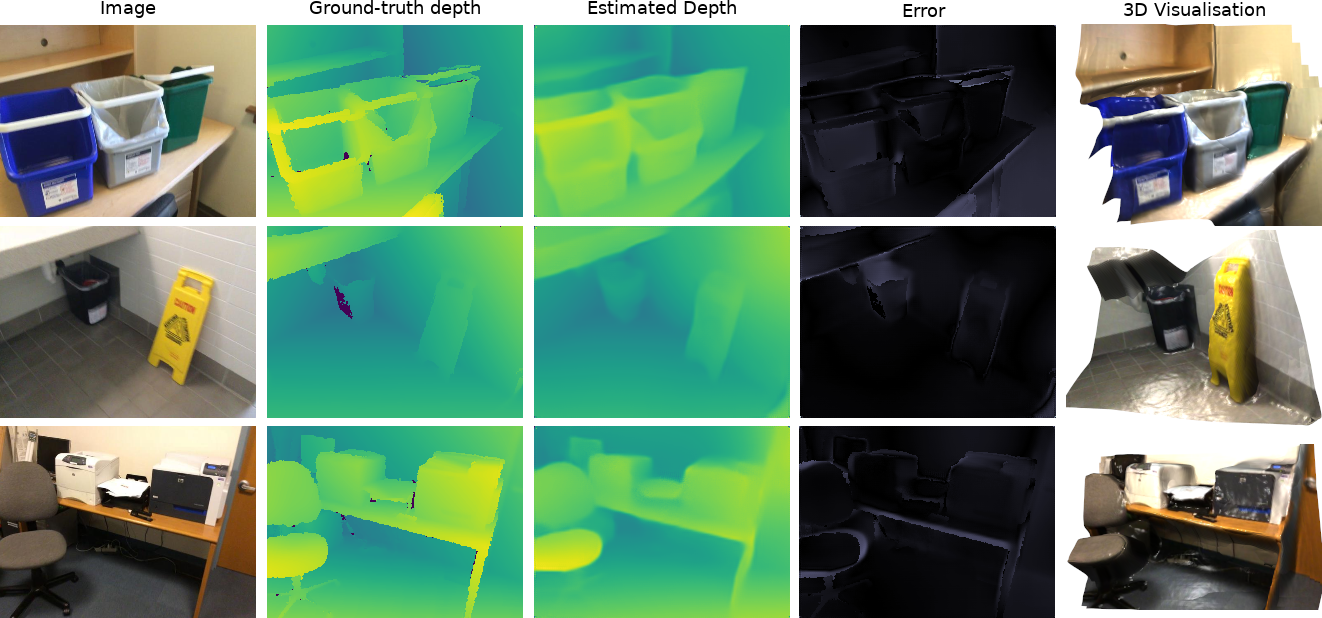}
\caption{Example reconstructions on selected scenes from the validation set of the ScanNet dataset. The specular lighting in the 3D visualisation has been exaggerated to better highlight structure. Best viewed on a computer screen.}
\vspace{2mm}\hrule
\label{fig:reconstr_qualitative}
\end{figure*}

We also visually present example reconstructions created by our system in Figure \ref{fig:reconstr_qualitative}.

\subsection{Trajectory Estimation}
We evaluate the Absolute Trajectory Error (ATE) of our proposed system and compare it against CNN-SLAM and CodeSLAM. For the sake of completeness, we also include comparison with DeepTAM, despite the fact that it does not achieve interactive (real-time) performance. We have used the same version of CodeSLAM that was used to generate the results in its respective paper. The code for CNN-SLAM is not available and the authors of DeepTAM do not provide a combined tracking and mapping system and we were not able to reproduce the results reported in the paper. For this reason, we include the numbers from the DeepTAM paper and evaluate our system on the same set of trajectories. We omit the room and plant sequences as they contain significant number of dropped frames, which skew the results. In order to limit computation we have disabled the geometric factor for the evaluation. 

The results are presented in Table~\ref{fig:ate_results}. We outperform CodeSLAM in all of the sequences and CNN-SLAM in all but one. In most cases we also achieve comparable results to DeepTAM, while still maintaining real-time performance.

\begin{table}
\centering
\def\arraystretch{1.0}
\setlength\tabcolsep{1.8pt}
\begin{tabular}{p{1.2cm} c c c c}
    \toprule
    \multicolumn{5}{c}{\textbf{Abs. Trajectory Error [m]}} \\
    \midrule
    Sequence & Ours & CNN-SLAM & DeepTAM\textbf{*} & CodeSLAM\textbf{*} \\ 
    \midrule
    fr1/360    & 0.142          & 0.500           & \textbf{0.116}  & 0.165 \\
    fr1/desk   & 0.119          & 0.095           & \textbf{0.078}  & 0.654 \\
    fr1/desk2  & 0.091          & 0.115           & \textbf{0.055}  & 0.181 \\
    fr1/rpy    & \textbf{0.047} & 0.261           & 0.052           & 0.078 \\
    fr1/xyz    & 0.064          & 0.206           & \textbf{0.054}  & 0.170 \\
    \bottomrule
\end{tabular}
\caption{Evaluation of our system on the validation sequences of the TUM RGB-D Benchmark\cite{Sturm:etal:IROS2012}.
We compare the ATE RMSE[m] against CNN-SLAM and DeepTAM, results for which have been taken from \cite{Zhou:etal:ECCV2018}.
CNN-SLAM was run without pose graph optimisation and our system without loop closures. Both CNN-SLAM and DeepTAM were initialised with the network from CNN-SLAM and we used our own initialisation. We've omitted the room and plant sequences as they contain significant frame drops that might skew the results.
\textbf{*Not real-time performance}}
\label{fig:ate_results}
\vspace{2mm}\hrule
\end{table}

\section{Performance and Implementation}
For a visual demonstration of the speed of the system please see the associated video, which contains real-time video recordings of our system in action. 

Our SLAM system has been implemented in C++ with the dense image warping, optimisation and camera tracking offloaded to GPU with CUDA, while the reprojection and the sparse geometric error factors are computed on the CPU. We run the network, CUDA kernels and visualization on a single NVIDIA GTX 1080 GPU and use image resolution of $256\times192$.

The network is ran on initialisation of each keyframe in order to obtain an initial code (depth) prediction and the Jacobian $\frac{\delta \matD}{\delta \vecc}$ that is used later in  optimisation. This requires around 340 ms, with only 16 ms of it spent on the forward pass of the network and the rest on calculating the Jacobian using \textit{tf.gradients}. This is due to the inefficiency of its backward-mode auto-differentiation based implementation which is optimised for gradients of scalar functions commonly used in machine learning. In our case, obtaining the derivative of each output pixel with respect to the latent representation requires a significant amount of passes through the network. This time can be drastically reduced with engineering effort and we predict it should be possible to reduce the overall time required to run the network to around 30 ms.

The incoming new camera images are tracked against the latest keyframe at around 250 Hz. Once the system initialises a next keyframe, we optimise the whole map representation in batch until convergence. The mapping steps are interleaved with  tracking the camera in order to keep up with new images.

The overall performance of the system varies greatly depending on the amount of connectivity between neighbouring keyframes specified by the user, the types of factors enabled, the number of factors relinearised within the iSAM2 algorithm and the occurrence of loop closures. With an explorative-type motion we achieve interactive real-time speeds, where we typically limit our system to only use the photometric and reprojection error to allow it to keep up with the fast camera movement. In a local reconstruction scenario like tabletop AR or room scale reconstruction where there is less exploration we can enable the geometric error to obtain better quality reconstructions. It is possible to further speed up the system performance through engineering effort which is part of planned future work.

The implementation of our system will be released on GitHub and shall be available under the following link: \href{https://github.com/jczarnowski/DeepFactors}{\textit{https://github.com/jczarnowski/DeepFactors}}


\section{Conclusions}

We have presented DeepFactors, a real-time probabilistic dense SLAM system built using the  the concept of learned compact depth map representation. We have demonstrated that our system achieves greater robustness and precision by combining different paradigms from classical SLAM with priors learned from data in a standard factor-graph probabilistic framework. The use of a standard framework allows it to be easily extended with different sensor modalities, which was not previously possible in the context of purely dense SLAM. An efficient C++ implementation and careful choices in the SLAM design enable real-time performance.

In future, we would like to explore the idea of including the structure-from-motion optimisation within the compact depth code training. This could allow obtaining a code manifold that is specifically trained to be later used in a mapping environment.
Moreover, learning the code representation in an unsupervised manner based on the intensity images only could be an interesting experiment. An inclusion of a relative-pose prediction network could also robustify the camera tracking.

We would also like to work on improving the performance of the current system, focusing on a faster method of obtaining the network Jacobian and a better GPU implementation of the geometric factor. 

\bibliography{IEEEabrv,robotvision}
\bibliographystyle{IEEEtran}

\end{document}

%% file: symbols.tex
\newcommand{\ignore}[1]{}

\newcommand{\ba}{\begin{array}}
\newcommand{\ea}{\end{array}}
\newcommand{\bc}{\begin{center}}
\newcommand{\ec}{\end{center}}
\newcommand{\be}{\begin{enumerate}}
\newcommand{\ee}{\end{enumerate}}
\newcommand{\bea}{\begin{eqnarray}}
\newcommand{\eea}{\end{eqnarray}}
\newcommand{\beas}{\begin{eqnarray*}}
\newcommand{\eeas}{\end{eqnarray*}}
\newcommand{\beq}{\begin{equation}}
\newcommand{\eeq}{\end{equation}}
\newcommand{\bfig}{\begin{figure}}
\newcommand{\efig}{\end{figure}}
\newcommand{\bi}{\begin{itemize}}
\newcommand{\ei}{\end{itemize}}
\newcommand{\bpic}{\begin{picture}}
\newcommand{\epic}{\end{picture}}
\newcommand{\btabular}{\begin{tabular}}
\newcommand{\etabular}{\end{tabular}}
\newcommand{\btable}{\begin{table}}
\newcommand{\etable}{\end{table}}

\newcommand{\es}{\vfill
                 \rule[-6mm]{170mm}{0.7mm} \\
                 \redw{{\tiny
		  \hfill S-\theslide}}
                 \end{slide}}

\newcommand{\matxx}[1]{{\mathtt #1}}
\newcommand{\vecXX}[1]{{\mathbf {#1}}}

\def \hbar {{\bar{h}}}

\def \vecc {{\vecXX{c}}}

\def \vecx {{\vecXX{x}}}
\def \vecy {{\vecXX{y}}}

\def \matD {{\matxx{D}}}

\def \matI {{\matxx{I}}}
\def \matJ {{\matxx{J}}}

\def \matT {{\matxx{T}}}

\makeatletter
\renewcommand*\env@matrix[1][*\c@MaxMatrixCols c]{%
  \hskip -\arraycolsep
  \let\@ifnextchar\new@ifnextchar
  \array{#1}}
\makeatother